\documentclass[letterpaper, 10 pt, conference]{ieeeconf}  

\IEEEoverridecommandlockouts                              

\usepackage{amsmath} 

\usepackage[dvipsnames,table,xcdraw]{xcolor} 
\usepackage{graphicx} 
\usepackage{multirow} 
\usepackage{etoolbox}
\makeatletter
\patchcmd{\@makecaption}
{\scshape}
{}
{}
{}
\makeatletter
\patchcmd{\@makecaption}
{\\}
{:\ }
{}
{}
\makeatother
\newcommand{\tablevspace}{\vspace{-2.5em}}

\usepackage{subcaption}

\newcommand{\figvspace}{\vspace{-2.0em}}

\def\secref#1{Sec.~\ref{#1}}
\def\figref#1{Fig.~\ref{#1}}
\def\tabref#1{Tab.~\ref{#1}}
\def\eqref#1{Eq.~(\ref{#1})}

\title{\LARGE \bf
Real-time LIDAR localization in natural and urban environments
}

\author{Georgi Tinchev, Adrian Penate-Sanchez and Maurice Fallon
	\thanks{The authors are with the Oxford Robotics Institute, University of Oxford, United Kingdom. 
	\texttt{\{gtinchev,adrian,mfallon\}@robots.ox.ac.uk}}%
	\thanks{This work was supported by EPSRC RAIN and ORCA Robotics Hubs (EP/R026084/1 and EP/R026173/1 respectively) 
	and EU H2020 project Memory of Motion (MEMMO, project ID: 780684). M. Fallon is supported by a Royal Society University 
	Research Fellowship.}
}

\begin{document}

\maketitle
\thispagestyle{empty}
\pagestyle{empty}

\begin{abstract}

Localization is a key challenge in many robotics applications. In this work we explore LIDAR-based global localization in both 
urban and natural environments and develop a method suitable for online application. Our approach leverages efficient deep 
learning architecture capable of learning compact point cloud descriptors directly from 3D data. The method uses an efficient 
feature space representation of a set of segmented point clouds to match between the current scene and the prior map. We 
show that down-sampling in the inner layers of the network can significantly reduce computation time without sacrificing 
performance. We present substantial evaluation of LIDAR-based global localization methods on nine scenarios from six datasets 
varying between urban, park, forest, and industrial environments. Part of which includes post-processed data from 30 
sequences of the Oxford RobotCar dataset, which we make publicly available. Our experiments demonstrate a factor of three 
reduction of computation, 70\% lower memory consumption with marginal loss in localization frequency. The proposed method 
allows the full pipeline to run on robots with limited computation payload such as drones, quadrupeds, and UGVs as it does not 
require a GPU at run time.

\end{abstract}

\section{INTRODUCTION}
\label{sec:intro}

Robotic systems are being developed in a large variety of environments. Those environments range from underwater 
survey~\cite{millerIJRR18,paullIJRR18}, aerial inspection~\cite{hunterICRA19}, indoor autonomy~\cite{vidriloDatasetIJRR15}, 
outdoor navigation~\cite{geiger2013vision,malagaDatasetIJRR}, and natural outdoor 
exploration~\cite{rosarioDatasetIJRR19,underwood2015lidar,garforth2019visual}. In all of these environments the ability to 
localize a robot against a prior map of some class is studied as it is one of the most basic and fundamental capabilities in 
robotics.

In our work we aim to achieve a localization performance which is robust across a broad range of environments. To obtain 
state-of-the-art performance in varied environments we take advantage of neural network architectures that can learn the most 
appropriate representation of the environment. In this paper we present an approach that learns features capable of generalizing 
between multiple environments.

When designing a solution to the localization problem, one needs to address the limited resources available on-board 
an autonomous robot. In contrast to other fields of engineering or computer science, the robot is mobile and needs to 
carry its own computation and power resources. For example,~\cite{resourceConstrainedIROS16} 
and~\cite{resourceConstraintsRAL17}  studied the need to co-design a mobile robot's software and 
hardware. While deep learning solutions have gained popularity due to their performance, it has been 
difficult to leverage them on smaller 
light-weight platforms because they frequently require GPU hardware that consumes large amounts of 
power. The solutions we aim for should be able to use the power of deep learning approaches while not imposing
the need for a GPU on the robot. This has clear advantages as it makes the system
more applicable to many different robotics platforms.


\begin{figure*}[t]
	\centering
		\includegraphics[trim=0 8 0 10, clip, width=\textwidth]{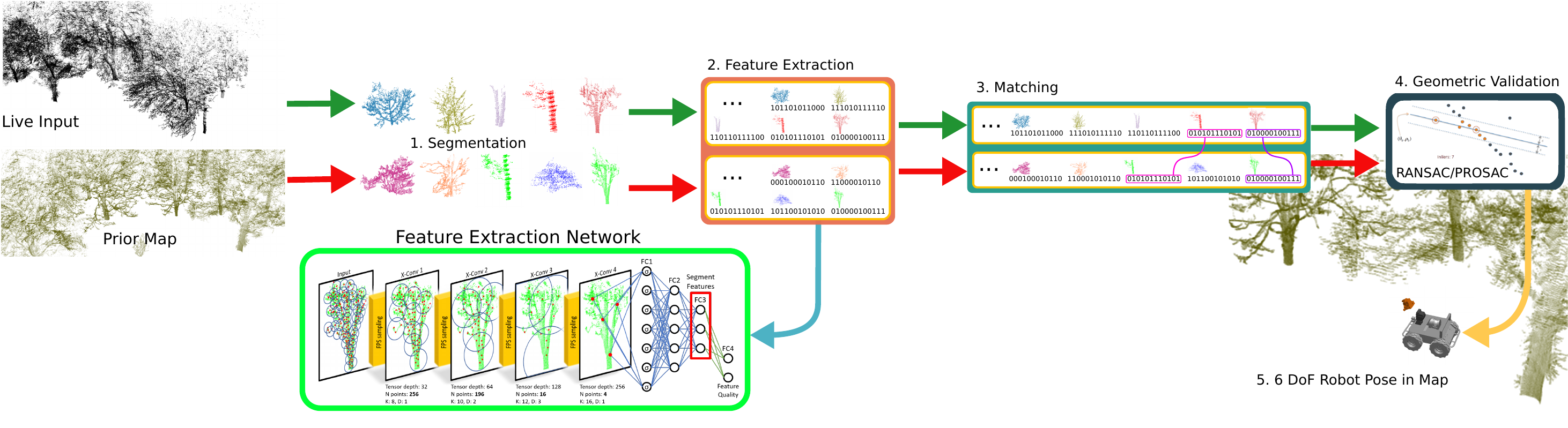}
	\caption{Pipeline outlining our approach to global pose estimation which localizes a live input point cloud within a prior map. 
	The feature extractor network is crucial for enabling efficient CPU performance.}
	\label{fig:teaser}
	\figvspace
\end{figure*}


In this paper we present an approach for real-time LIDAR segment-based localization, shown in~\figref{fig:teaser}. This new 
method is three times faster than the previous best~\cite{tinchev2019learning} with marginal performance loss. Another 
contribution of this paper is the creation of rich annotations for LIDAR data for validation involving over $50\,\text{km}$ of 
testing data and $250\,\text{km}$ of training data. For this, we have aligned and annotated over $2.5$ million point cloud 
segment matches from the Oxford RobotCar dataset~\cite{RobotCarDatasetIJRR}. This substantial effort on improving the 
dataset will be made available publicly and is one of the biggest LIDAR annotated datasets. We also introduce a second, smaller 
dataset on which we run field trials on the legged robot ANYmal~\cite{ANYmal_AR17}.

In summary, the contributions of our work are as follows:

\begin{itemize}
	\item We propose a more efficient method for LIDAR localization achieving performance equivalent to 
	state-of-the-art GPU-dependent approaches without the need for heavy computational resources on the robot.
	By introducing a down-sampled representation in each of the layers in the neural network. The neural network 
	shifts from a representation where all points are maintained throughout the architecture to a novel model 
	that performs down-sampling of the points to limit the computational complexity during inference. This 
	increases the speed of our method by a factor of three and reduces memory consumption by 70\%.
	\item An efficient CPU implementation to solve the Furthest Point Sampling algorithm that achieves a speedup of 
	up to $80$\,x, allowing this technique to be applied within the neural network in a CPU in order to handle the
	efficient down-sampling of the point cloud within the neural network when using only on a CPU.
	\item We present extensive evaluation of our new approach by comparing to four baselines on six different 
	datasets. In total this represents more than $50\,\text{km}$ of testing data in a variety of environments 
	- park, forest, urban, and industrial scenery. Furthermore, we evaluate the generalization abilities of our approach given 
	different training and testing environments, sensor modalities with different frequency of operation, angular resolution, and 
	mode of operation (2D LIDAR vs 3D LIDAR).
	\item As a final contribution we offer the community access to the annotated real-world LIDAR dataset. This has 
	been possible by carefully aligning the LIDAR data from 30 different $10\,\text{km}$ trials from the Oxford 
	RobotCar dataset. These trials were 
	collected throughout the whole year. This allows us to retrieve point correspondences between different 
	experiences with a very high degree of accuracy. There are multiple LIDAR samples of the same object, 
	tree or	house which is beneficial for supervised neural network training in the presence of appearance 
	change,	viewpoint variation, and other challenges.
\end{itemize}

In the following sections we present the literature relevant to segment-based LIDAR localization.~\secref{sec:methodology} 
describes the proposed network architecture, while~\secref{sec:data} describes the used datasets and the proposed 
improvements of the Aligned Oxford RobotCar dataset. In~\secref{sec:results} we present extensive evaluation of all LIDAR 
segment-based methods in terms of feature performance, generalization ability, and computational complexity.

\section{Related Work}
\label{sec:related}

The research presented in this paper focuses on global localization (or loop closure within a SLAM context) using clusters of 
individual LIDAR segments. This segment-based approach has been shown to be reliable for robot 
localization~\cite{original_segment_paper2012,dub2017icra} and has the advantage of being physically meaningful. In this 
section the related work will contextualize our approach by describing segment-based approaches using LIDAR data.

A recent approach for global LIDAR localization is to extract small point cloud clusters or~\textit{segments} from 
the raw input point clouds which are then matched against a prior map and used to localize. A~\textit{segment} is a cluster of 
LIDAR points broadly but not directly comparable to an object. It might correspond to a patch of a flat 
wall, a vehicle or a tree trunk. This approach is used instead of determining local keypoints or directly registering the 
raw points individually.

\begin{figure*}[t!]
	\centering
	\includegraphics[trim=0 15 0 0, clip, width=\textwidth]{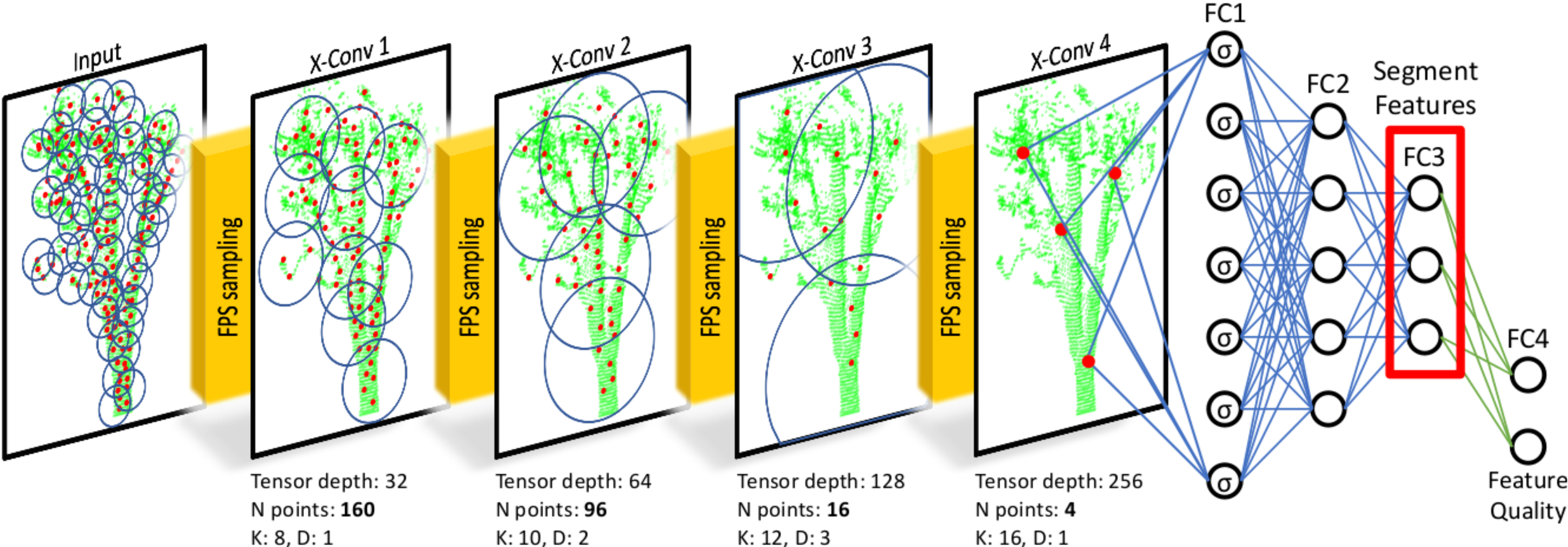}
	\caption{ The proposed network architecture. The input to the network is a point cloud segment of 256 points. At each layer 
		the number of points (N), nearest neighbors (K) and dilation rate (D) are specified. The descriptor of the network is 
		extracted 
		from the third fully connected layer (FC3). The architecture also has an additional classification branch (FC4), defining the 
		quality of the descriptor. Between each layer we utilize Farthest Point Sampling (FPS) to efficiently decrease the number of 
		points seen by the network at later stages.}
	\label{fig:network_architecture}
	\figvspace
\end{figure*}


This approach was first proposed by~\cite{original_segment_paper2012} where segments constituted the 
middle ground between local and global approaches. Global approaches can suffer if they are not robust to 
occlusions or change in the environment, but they can encapsulate the context of the landmarks/objects present in the 
scene. Conversely, local methods, similar to the works of~\cite{zhuang20123} and~\cite{steder2011place},  are designed to be 
robust to occlusions and change, but because they only characterize local regions they lose much of the contextual information. 
Segments, however, are large enough that they contain reliable and repeatable features while also being robust to occlusions 
and temporal changes.

Segment localization were first applied to LIDAR localization in the works of~\cite{dub2017icra,segmap2018,segmapIJRR} and we 
draw a lot of inspiration from these works. The SegMatch approach extracted more reliable feature matches than the then 
state-of-the-art 3D point descriptors and could achieve a large number of robust localizations. Similarly to~\cite{zachCVPR15}, 
SegMatch's localizations are obtained by a geometric match pruning of the proposed segments. The centroids of a set of 
segments are used as input to RANSAC (\cite{fischler1981random}) which produces the final pose estimate. 
In~\cite{segmap2018} segment localization evolved into a segment-based SLAM that introduced a new segment feature that also 
encoded semantics and volumetric shape. This novel approach learned features from a voxelized input space that were 
suitable for localization and semantic description of the map. The approach was evaluated in a realistic urban setting 
and in industrial buildings. In our previous work~\cite{tinchev18seeing,tinchev2019learning} we proposed a novel descriptor 
capable of generalizing to natural environments and a new network architecture suitable for CPU operation at test time. 

In this work we propose new neural network architecture that performs down-sampling of the point cloud between layers 
to create a clustering representation within the network instead of per point representation. The main challenge down-sampling
within the neural network poses from a computational standpoint is that such a computation can be easily handled with a GPU 
but requires an efficient CPU implementation that allows it to run in real time, which we describe in this paper. By using 
this new neural network we can reduce the required computation by a factor of three and occupy 70\% less memory at a 
marginal performance cost.

\section{METHODOLOGY}
\label{sec:methodology}

The main focus of this work is on global localization where a scene point cloud is matched against a prior map. The localization 
pipeline is outlined in~\figref{fig:teaser}. In this section we describe our proposed method, named Down-sampled Segment 
Matching (\textbf{DSM}), which has the following properties: 1) a more efficient neural network architecture for point cloud 
description suitable for both GPU and CPU inference, and 2) a down-sampled segment descriptor at 
each layer, which achieves 3) three times quicker inference and 70\% less memory utilization than the previous 
best~\cite{tinchev2019learning}. The reason for this is that at each layer of the model presented in~\cite{tinchev2019learning}, 
the 
$\mathcal{X}$-conv operator~\cite{li2018pointcnn} computes the k-nearest neighbors for each point, which is very expensive. 
Here,
by carefully sampling the points, we reduce the required computation by a factor of three, 
while retaining the important information for segment matching. In the following sections 
we describe the architecture and down-sampling in more detail.

\subsection*{Network Architecture}

The architecture for \textbf{DSM} is presented in Figure~\ref{fig:network_architecture}. We have opted for a 
smaller point size per layer, in order to reduce the computation load in comparison to~\cite{tinchev2019learning}. The input to 
our network is a single segment of 256 points. These points are then down-sampled to $160$ in the first layer with 
the number of nearest neighbours $N=8$. A dilation of $1$ is applied, resulting in the layer processing every point from the 
input point cloud. Intuitively, our first layer reduces the input by 40\%. The second layer of the model has a point size of $96$ 
with the number of nearest neighbors, $N=10$ and dilation of 2. In this manner we ensure that every point in the new layer is 
connected to $10$ neighbors from the layer above, which are equally spaced, every 2 points. Therefore, a single point 
aggregates features in a hierarchical way. Similarly, our third layer has a point size of $16$, each point is connected to $N=12$ 
neighbors, dilated every $3$ points. This representation shrinks the number of points by 80\% with each remaining point being 
much more informative. The final $\mathcal{X}$-conv layer has size $4$, with $N=16$, and dilation of $1$. This layer ensures 
that every point in the last layer \textit{sees} all the possible information, while retaining a small number of points. We have 
empirically evaluated that this architecture consumes 70\% less memory for the same input batch, when compared to our 
previous work~\cite{tinchev2019learning}, and has slightly fewer parameters ($280$\,K compared to $300$\,K). We also 
proposed a different selection methods for points in each subsequent layer including random sampling, inverse density 
sampling, and Farthest Point Sampling (FPS). Our experiments showed that FPS improves the overall results, as it selects points 
based on coverage. Using this evidence we established solid grounds on which to propose our novel neural network architecture 
(\textbf{DSM}) that reduces the computational cost greatly compared to the most computationally efficient 
baseline~\cite{tinchev2019learning}.


\begin{figure*}[t]
	\centering
	\includegraphics[trim=0 80 0 0, clip, width=\textwidth]{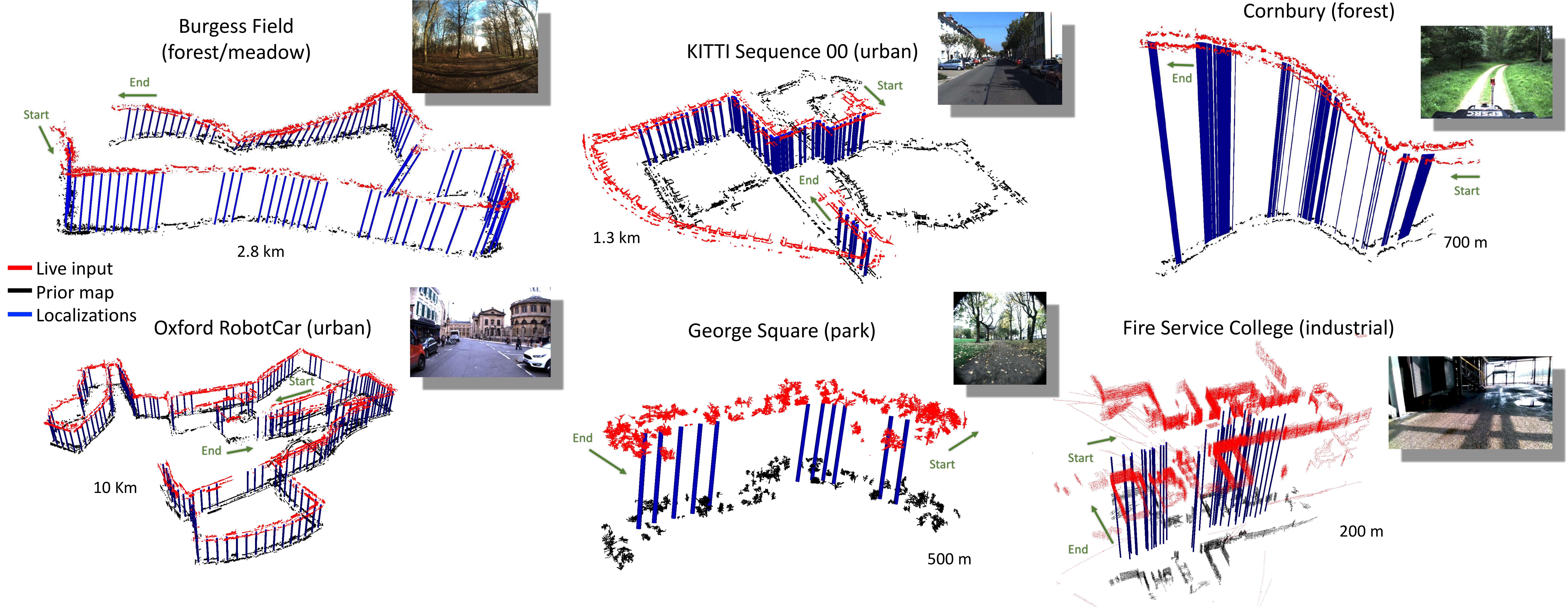}
	\caption{Top-down visual preview of all the datasets considered in our work. The sequence of live 
		point clouds are denoted in red and have been vertically offset from the map clouds (in black) in $Z$ 
		for visualization purposes. The starting and ending poses for each dataset are shown with the 
		indicated green arrows denoting the traversed trajectory. The vertical blue lines indicate a correct 
		place recognition in the map's reference frame. The length of each dataset and a representative 
		camera image are shown next to each dataset run.
	}
	\label{fig:combined_datasets}
	\figvspace
\end{figure*}


\subsection*{Efficient sampling in compact neural architectures}

This section focuses on Farthest Point Sampling (FPS) and describes an efficient implementation 
of FPS suitable for CPU execution during inference.

The input to the sampling approach are the global $x,y,z$ coordinates of all points, and the number 
of sampled points required. The first step of the sampling procedure is to select a random point and 
compute the distance from that point to the rest of the points in the segment. The first point is 
added to the set of visited points. Thereafter, iteratively the algorithm selects the point that has the 
biggest distance between its previous point and the rest of the unvisited points and adds it to the list of 
visited points. This continues until the stopping criteria is reached, i.e the cardinality of the visited set is 
equal to the input parameter. In this manner the selected points that we convolve are separated as far apart as possible. This 
helps the convolution retain similar amount of information but with only a small fraction of the points.

This algorithm has been efficiently parallelized using CUDA on a GPU~\cite{qi2017pointnetplusplus}. The authors present results 
on the effect of randomness when choosing the first point. When training our model we still use the GPU as much as possible as 
this computation happens offline but for test time we resort to our efficient CPU implementation, which we detail below.

In our work, we wish to utilize sampling in each layer of our architecture while being able to run on a CPU. Initially, we developed 
our approach to iteratively choose the most distant point in each segment individually. This per-segment approach is not 
optimal, as it requires $O(PS)$ time complexity, where $P$ denotes the number of segments in a batch, while $S$ -- the number 
of selected points to sample. We improved this by implementing a sampling technique directly on batches of segments in a 
tensor simultaneously. The data in each layer of our model is structured as a tensor, where the height of the tensor holds each 
point from the segment, the width corresponds to each of the points' dimensions (usually $x,y,z$), and the length represents 
the batch of segments in a point cloud. This way we can efficiently compute the distance matrix of the entire tensor, along the 
length of the tensor, and simultaneously select the farthest point in all segments.

While the above approach still needs to compute the distance matrix for each following farthest 
point, it is still more efficient than running on per-segment basis -- $O(S)$. In comparison to a 
per-segment approach, we gain $80$\,x speed-up, while compared to the thousands of 
cores that CUDA utilizes, we are just $30$\,x slower, but only using a CPU. By paying attention to the computational 
details of the sampling, our method operates at a frequency of $5\,\text{Hz}$ on a CPU.

\section{Data Structure}
\label{sec:data}

\begin{figure*}[t]
	\centering
	\begin{subfigure}[b]{0.45\textwidth}
		\includegraphics[width=\textwidth]{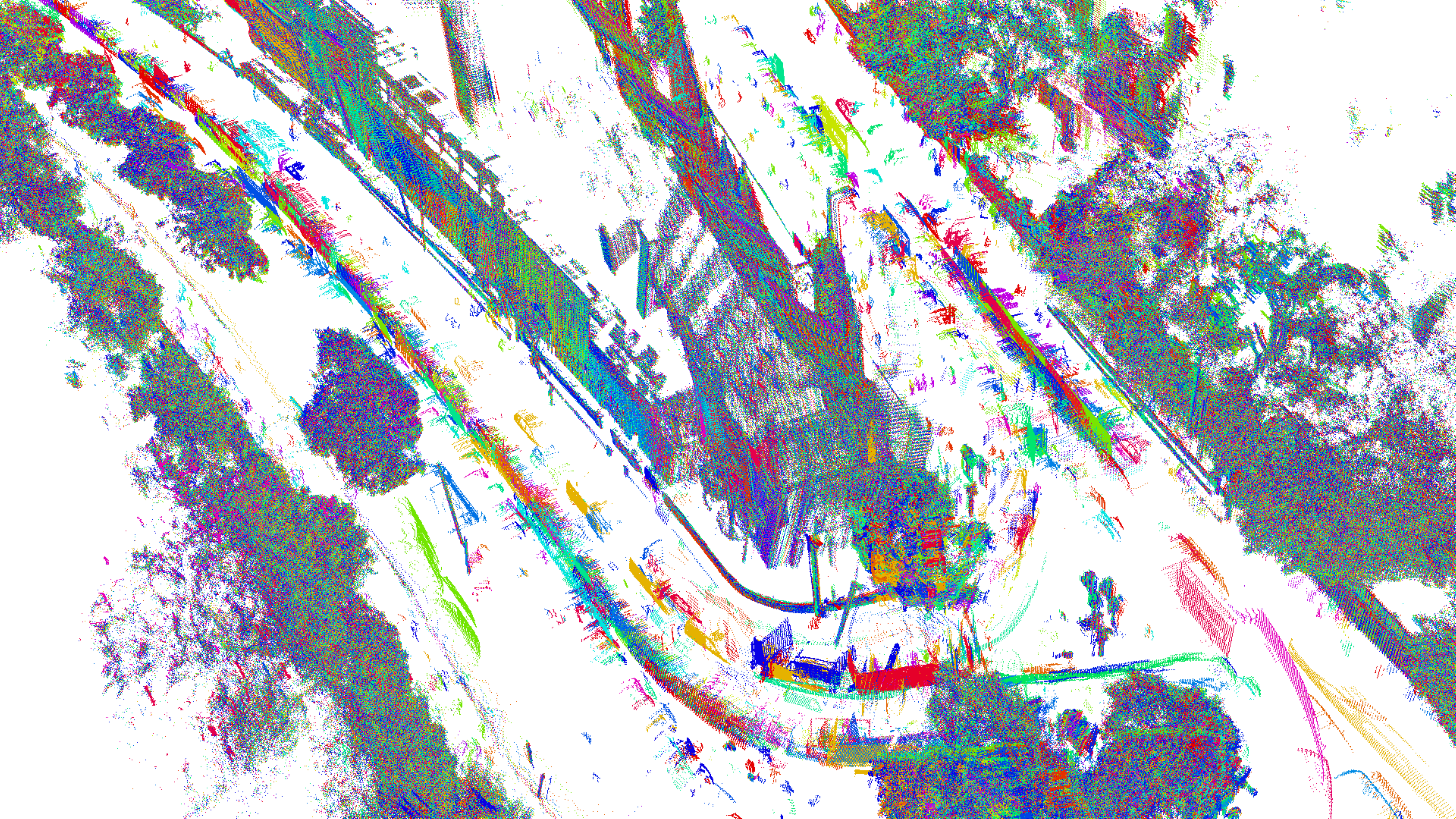}
	\end{subfigure}
	~
	\begin{subfigure}[b]{0.45\textwidth}
		\includegraphics[width=\textwidth]{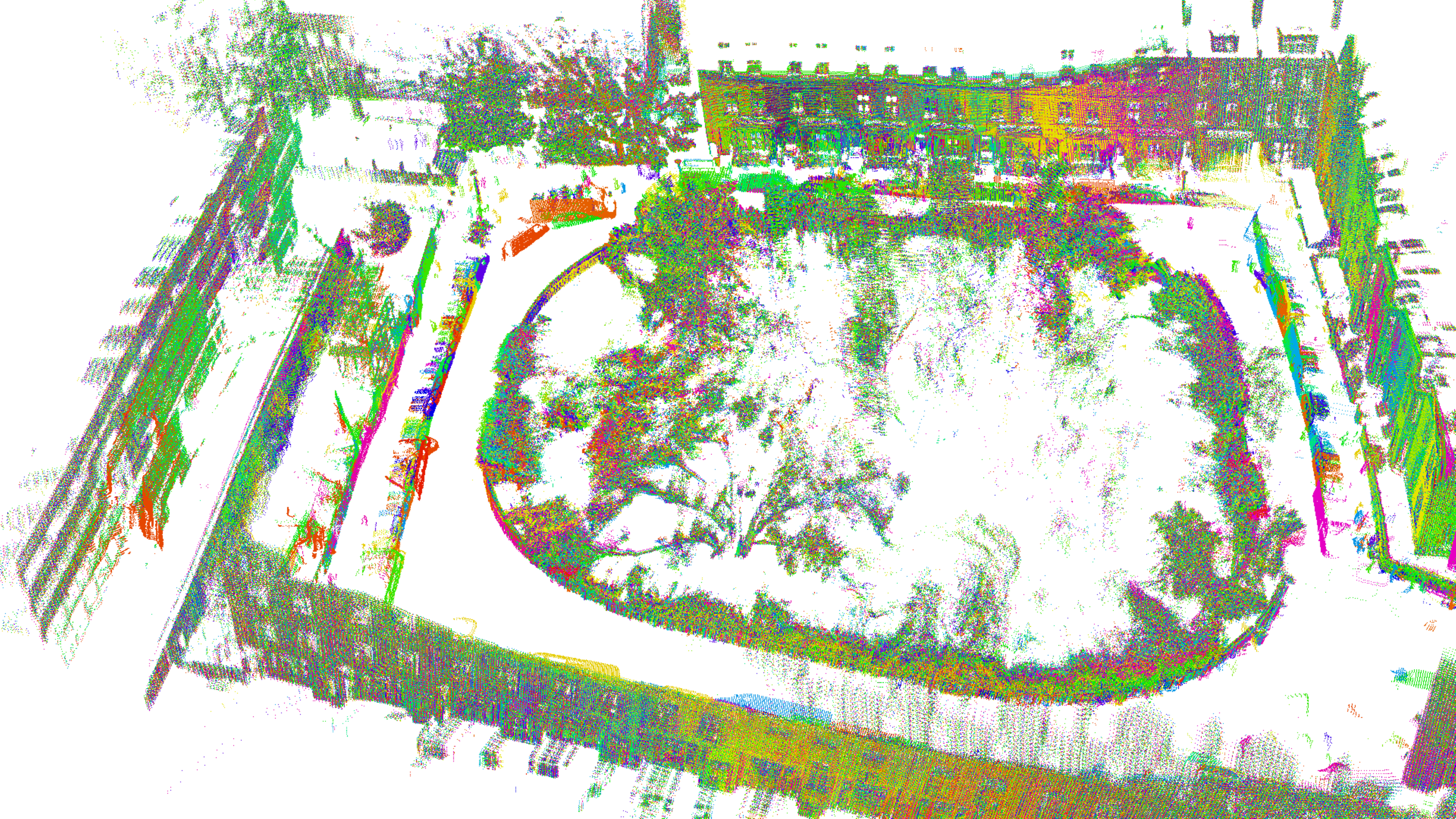}
	\end{subfigure} 
	\caption{Preview of the aligned Oxford RobotCar dataset across multiple sequences. Each of the 30 point clouds shown in the 
		figure is colored based on the scenario it has been generated from. Parks road junction (left) shows the precise alignment 
		of 
		buildings and vegetation. Wellington Square (right) illustrates a loop around a small park in Oxford, UK.}
	\label{fig:robotcar_dataset}
	\figvspace
\end{figure*}

A special effort has been made to present broad experimental results for our new method as well as other state-of-the-art 
segment-based localization methods~\cite{dub2017icra,segmap2018,tinchev18seeing,tinchev2019learning}. To provide the basis 
of our analysis we used six different datasets, collected using five different platforms in urban, industrial, and natural 
environments. A top-down view of the localizations at each environment is shown in~\figref{fig:combined_datasets}, while a 
video preview of our algorithm running on the datasets is available in the supplementary material. Nearly 320 kilometers of data 
has been analyzed in the results section. We believe that this level of testing is necessary to properly measure and analyze these 
methods. Of particular note is our large-scale analysis of the RobotCar dataset~\cite{RobotCarDatasetIJRR}, shown 
in~\figref{fig:robotcar_dataset}. As a contribution of this paper we make available an improved version of the dataset by 
providing refined ground truth poses of its 300 kilometers.

\paragraph*{Aligned Oxford RobotCar Dataset}
\label{sec:robotcar}

The Oxford RobotCar Dataset contains over 100 sequences of a single consistent route through Oxford, UK, captured over a 
period of a year on a vehicle equipped with various navigation sensors including a push-broom LIDAR. The dataset 
contains many different combinations of weather, traffic, and pedestrians. Data were collected in all weather conditions, 
including heavy rain, night time, direct sunlight, and snow. Road and building works over the period of a year significantly 
changed sections of the route throughout the full period.


\paragraph*{Ground Truth Annotation}

The existing Oxford RobotCar Dataset provides position estimates using a Global Positioning System (GPS) and Inertial 
Navigation System (INS) which can be used as a proxy for ground truth. However, for deep learning the object instances in the 
LIDAR data need to be tightly associated to one another. The error due to GPS is in many cases too large to be satisfactory for 
this. For this reason we employed two months of work annotating a more accurate ground truth. This improved ground truth 
allowed us to establish correspondences between LIDAR clouds to the accuracy needed for supervised learning. Two example 
scenes from the aligned dataset are shown in~\figref{fig:robotcar_dataset}. For a full preview of the 
improved dataset, please see the videos in the supplementary material.

The ground truth was created as follows: visual odometry provided a backbone with visual loop closures established within and 
across the 100 sequences~\cite{ChurchillIJRR2013}. This created an initial map that we then manually corrected by verifying 
LIDAR consistency and adding individual loop closures where necessary. This provides several samples where each object is seen 
and allows the feature learning to adapt to viewpoint changes and environment alterations.

\paragraph*{Data Organization}

In this work we have trained all models and baselines with 26 of the sequences, captured between June 2014 and October 
2015, containing a total of 140,287 samples from 59,503 classes. For testing we have used four sequences from November 2015 
that were not used during training. We have established 101,502 matching and 2,208,008 non-matching segments. Two 
segments were considered a match if the distance between their centroids is less than $0.5$\,m; and a non-match when the 
distance is larger than $20$\,m. For localization, we generated 3D swathes from the push-broom sensor by incorporating the 
motion of the vehicle for every $30$\,m of distance traveled. This resulted in a total of 1,118 point clouds across all four testing 
sequences. 

We make the ground truth publicly available as a contribution of this paper. It has been added to the same web site as the 
original dataset\footnote[1]{https://robotcar-dataset.robots.ox.ac.uk/lidar/}. We believe this is very valuable as such 
a detailed annotation is difficult to attain. We hope it can be useful to other deep learning techniques such as feature learning 
and change detection. 

\section{RESULTS}
\label{sec:results}

\begin{figure*}[t!]
	\centering
	\begin{tabular}{@{}c@{}}
		\includegraphics[width=0.32\textwidth]{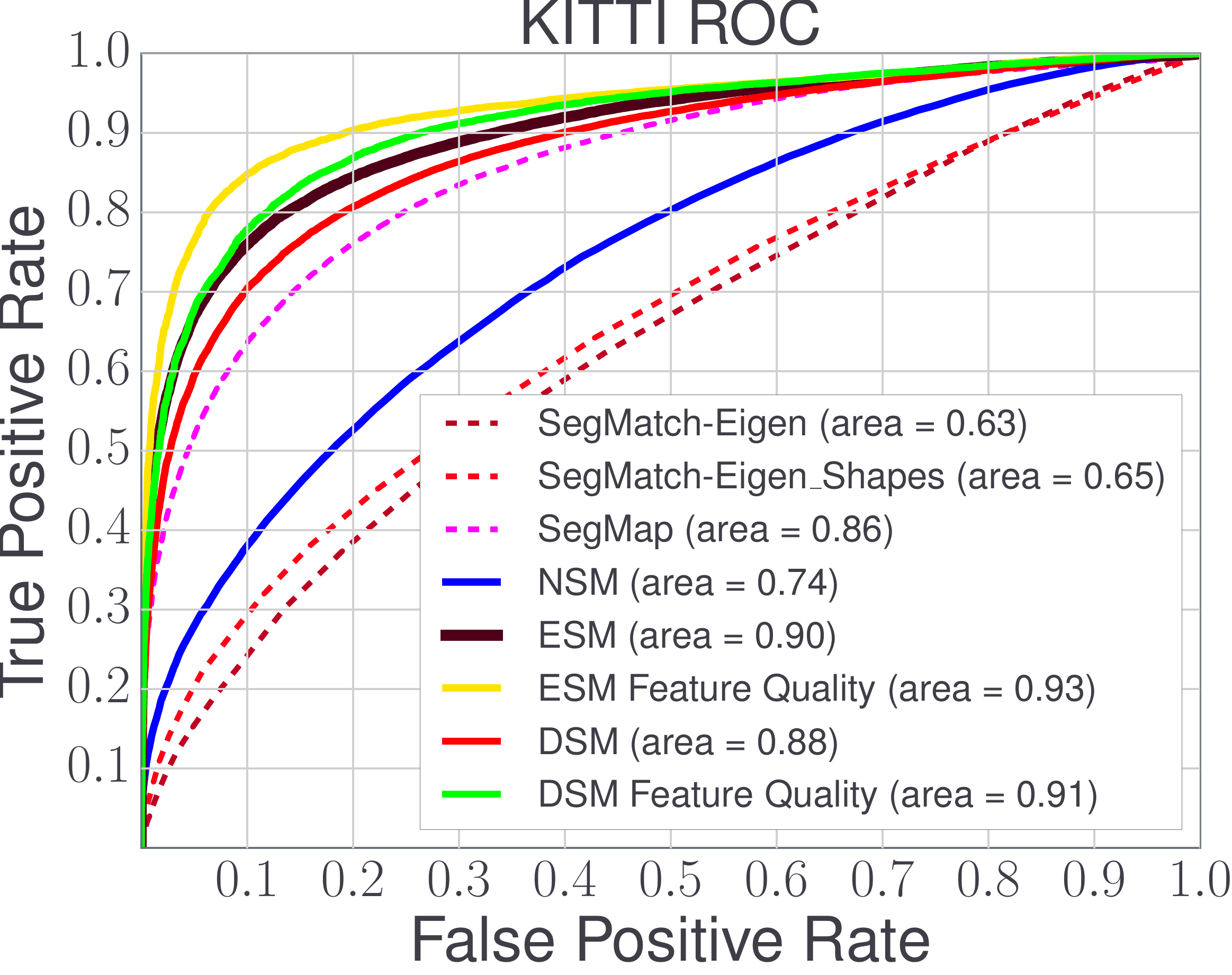}
	\end{tabular}
	~
	\begin{tabular}{@{}c@{}}
		\includegraphics[width=0.32\textwidth]{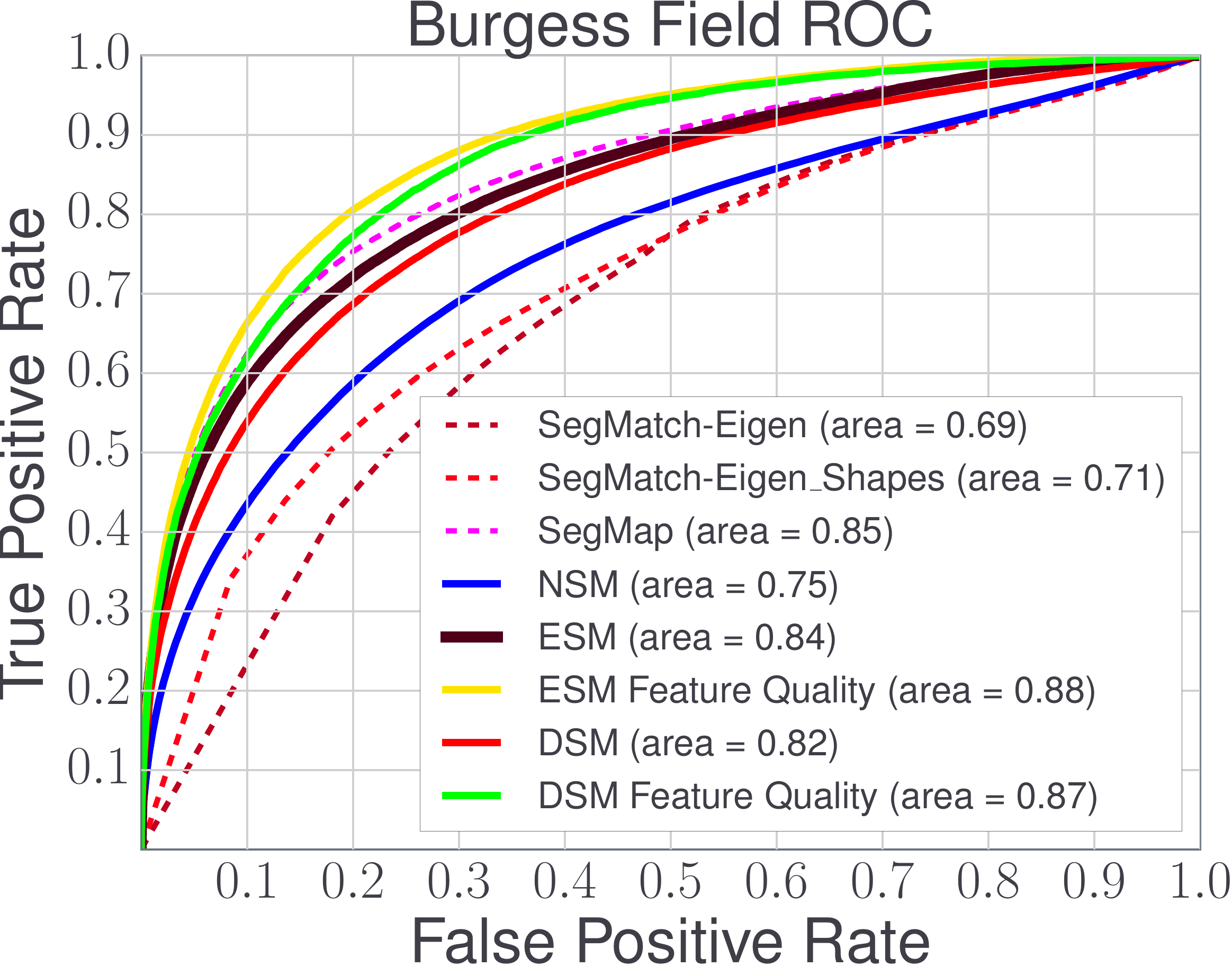}
	\end{tabular}
	~
	\begin{tabular}{@{}c@{}}
		\includegraphics[width=0.32\textwidth]{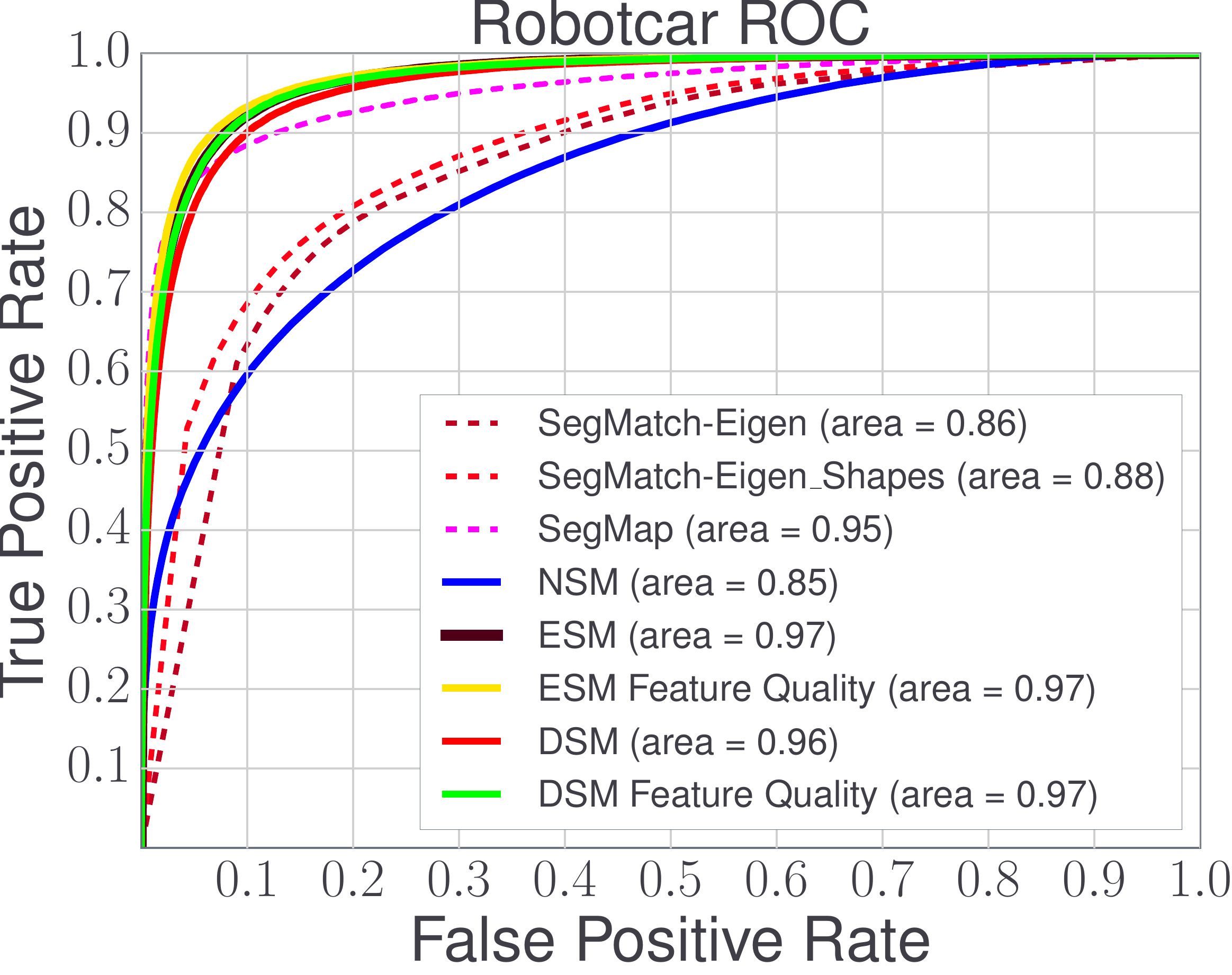}
	\end{tabular}
	\caption{Receiver Operating Characteristic (ROC) curves for the proposed method, \textbf{DSM}, and baseline approaches 
		on KITTI (urban, left), Burgess Field (natural, middle) and Oxford RobotCar (urban, right) datasets.}
	\label{fig:roc}
	\vspace{-1em}
\end{figure*}

\begin{table*}[t!]
	\resizebox{\textwidth}{!}{%
		\begin{tabular}{|cc|c|c|c|c|c|c||c|c|c|c|c|c|c|c|c|c|}
			\hline
			&  
			& \multicolumn{2}{c|}{\cellcolor[HTML]{FFFC9E}\textbf{SM-Eig}} 
			& \multicolumn{2}{c|}{\cellcolor[HTML]{FFFC9E}\textbf{SM-ESF}} 
			& \multicolumn{2}{c||}{\cellcolor[HTML]{FFFC9E}\textbf{NSM-RF}} 
			& \multicolumn{2}{c|}{\cellcolor[HTML]{FFFC9E}\textbf{SegMap}} 
			& \multicolumn{2}{c|}{\cellcolor[HTML]{FFFC9E}\textbf{ESM}} 
			& \multicolumn{2}{c|}{\cellcolor[HTML]{FFFC9E}\textbf{ESM-FQ}} 
			& \multicolumn{2}{c|}{\cellcolor[HTML]{FFFC9E}\textbf{DSM}} 
			& \multicolumn{2}{c|}{\cellcolor[HTML]{FFFC9E}\textbf{DSM-FQ}} 
			\\ 
			\hline
			\rowcolor[HTML]{FFCE93} 
			\multicolumn{1}{|c|}{\cellcolor[HTML]{E8FFC7}\textbf{Test}} 
			& \cellcolor[HTML]{E8FFC7}\textbf{Train} & \textbf{Area} & \textbf{Perc.} 
			& \textbf{Area} & \textbf{Perc.} & \textbf{Area} 
			& \textbf{Perc.} & \textbf{Area} & \textbf{Perc.} & \textbf{Area} 
			& \textbf{Perc.} & \textbf{Area} & \textbf{Perc.} & \textbf{Area} 
			& \textbf{Perc.} & \textbf{Area} & \textbf{Perc.} 
			\\ 
			\hline \hline
			\rowcolor[HTML]{EFEFEF} 
			\multicolumn{1}{|c|}{\cellcolor[HTML]{DAE8FC}} & \cellcolor[HTML]{DAE8FC}\textbf{KITTI} 
			& 0.632 & 68\% & 0.655 & 70\% & 0.738 & 79\% & 0.858 & 92\% & 0.902 & 97\% 
			& {\color[HTML]{00A600} \textbf{0.930}} & {\color[HTML]{00A600} \textbf{100\%}} 
			& 0.880 & 95\% & 0.911 & 98\% 
			\\ \cline{2-18} 
			\multicolumn{1}{|c|}{\cellcolor[HTML]{DAE8FC}} & \cellcolor[HTML]{FFCCC9}\textbf{Burgess Field} 
			& 0.666 & 72\% & 0.711 & 76\% & 0.736 & 79\% & 0.764 & 82\% & 0.828 & 89\% & 0.888 & 95\% 
			& 
			0.864 
			& 93\% & {\color[HTML]{F45D01} 0.920} & {\color[HTML]{F45D01} 99\%} \\ \cline{2-18} 
			\multicolumn{1}{|c|}{\multirow{-3}{*}{\cellcolor[HTML]{DAE8FC}\textbf{KITTI}}} 
			& \cellcolor[HTML]{9AFF99}\textbf{RobotCar} & 0.680 & 73\% & 0.709 & 76\% & 0.752
			& 81\% & 0.822 & 88\% & 0.873 & 94\% & 0.890 & 96\% & 0.807 & 87\% & 0.825 & 89\% 
			\\ \hline \\[-0.5em] \hline 
			\rowcolor[HTML]{EFEFEF} 
			\multicolumn{1}{|c|}{\cellcolor[HTML]{FFCCC9}} & \cellcolor[HTML]{FFCCC9}\textbf{Burgess Field} 
			& 0.685 & 78\% & 0.714 & 81\% & 0.753 & 85\% & 0.850 & 96\% & 0.836 & 95\% 
			& {\color[HTML]{00A600} \textbf{0.883}} & {\color[HTML]{00A600} \textbf{100\%}} 
			& 0.815 & 92\% & {\color[HTML]{F45D01} 0.869} & {\color[HTML]{F45D01} 98\%} 
			\\ \cline{2-18} 
			\multicolumn{1}{|c|}{\cellcolor[HTML]{FFCCC9}} & \cellcolor[HTML]{DAE8FC}\textbf{KITTI} 
			& 0.626 & 71\% & 0.697 & 79\% & 0.713 & 81\% & 0.781 & 88\% & 0.817 & 93\% & 0.868 
			& 98\% & 0.797 & 90\% & 0.849 & 96\% 
			\\ \cline{2-18} 
			\multicolumn{1}{|c|}{\multirow{-3}{*}{\cellcolor[HTML]{FFCCC9}\textbf{Burgess Field}}} 
			& \cellcolor[HTML]{9AFF99}\textbf{RobotCar} & 0.710 & 80\% & 0.730 & 83\% & 0.732
			& 83\% & 0.808 & 92\% &  0.814 & 92\% & 0.853 & 97\% & 0.794 & 90\% & 0.839 & 95\% 
			\\ \hline \\[-0.5em] \hline
			\rowcolor[HTML]{EFEFEF} 
			\multicolumn{1}{|c|}{\cellcolor[HTML]{9AFF99}} & \cellcolor[HTML]{9AFF99}\textbf{RobotCar} 
			& 0.856 & 88\% & 0.880 & 91\% & 0.848 & 87\% & 0.953 & 98\% & {\color[HTML]{F45D01} 
				0.969} 
			& {\color[HTML]{F45D01} 99\%} & {\color[HTML]{00A600} \textbf{0.972}} 
			& {\color[HTML]{00A600} \textbf{100\%}} & {\color[HTML]{F45D01} 0.961} & {\color[HTML]{F45D01} 99\%} & 
			{\color[HTML]{F45D01} 
				0.967} 
			& 
			{\color[HTML]{F45D01} 99\%} 
			\\ \cline{2-18} 
			\multicolumn{1}{|c|}{\cellcolor[HTML]{9AFF99}} & \cellcolor[HTML]{DAE8FC}\textbf{KITTI} 
			& 0.760 & 78\% & 0.821 & 84\% & 0.815 & 84\% & 0.952 & 98\% & 0.934 & 96\% & 0.926 
			& 95\% & 0.928 & 95\% & 0.929 & 96\% \\ \cline{2-18} 
			\multicolumn{1}{|c|}{\multirow{-3}{*}{\cellcolor[HTML]{9AFF99}\textbf{RobotCar}}} 
			& \cellcolor[HTML]{FFCCC9}\textbf{Burgess Field} & 0.819 & 84\% & 0.853 & 88\% & 0.865 
			& 89\% 
			& 0.951 & 98\% & 0.934 & 96\% & 0.948 & 98\% & 0.922 & 95\% & 0.938 & 97\% \\ \hline
		\end{tabular}%
	}
	\caption{Generalization results - difference in performance when training data environment differs from testing data. The best 
	performing algorithm is colored green and the second best orange. \textit{Area} is the area below the ROC curve, while 
	\textit{Perc} is the performance relative to the best performing algorithm.}
	\label{table:percentage_results}
	\tablevspace
\end{table*}

The main focus of this work is to localize a 3D \textit{scene} point cloud with respect to a prior map. In this section we evaluate 
the individual components of our system. We also overview several baseline approaches before comparing performance
using six challenging datasets. 

We compare our method against two popular methods for segment-based localization - SegMatch\footnote[2]{\label{fn:impl}We 
use the open-source implementations of SegMatch/SegMap at https://github.com/ethz-asl/segmap.}~\cite{dub2017icra} and the 
data-driven approach SegMap\footnotemark[2]~\cite{segmap2018}, as well as our previous works - NSM~\cite{tinchev18seeing} 
and ESM~\cite{tinchev2019learning}. When evaluating our experiments we define the following abbreviations (ordered by 
publication date):

\begin{itemize}
	\item SM-Eig - SegMatch~\cite{dub2017icra} with only Eigenvalue features.
	\item SM-ESF - SegMatch~\cite{dub2017icra} using both Eigenvalue and Ensemble of Shape Histogram features (ESF).
	\item NSM-RF - NSM~\cite{tinchev18seeing} with Gestalt features and a Random Forest.
	\item SegMap~\cite{segmap2018} with voxel-based learning architecture.
	\item ESM and ESM-FQ\cite{tinchev2019learning} - baseline approach with and without a feature quality (FQ) classification 
	branch.
	\item \textbf{DSM} and \textbf{DSM-FQ} - The proposed method with and without the feature quality of the classification 
	branch.
\end{itemize}

Due to the modularity of our approach, we can evaluate each component of our system individually and compare it to 
the baseline methods. First, we examine the effect of the learned features and their performance on three different datasets - 
two urban scenarios and one parkland scenario. Second, we study the generalization ability of each method by looking at 
different permutations of training and testing data across the datasets. Third, we address a common issue faced by some 
approaches relating to rotation variation and viewpoint change. Finally, using the optimal set of system components, we analyze 
the overall execution of our system in terms of localization performance and runtime efficiency.

\subsection*{Feature Performance}

We begin the evaluation by looking at the descriptive power of the features. We chose the 
KITTI~\cite{geiger2013vision}, Burgess Field (own), and RobotCar~\cite{RobotCarDatasetIJRR} datasets, shown 
in~\figref{fig:combined_datasets}. We chose these three datasets due to the large amounts of labeled data and the different 
point cloud density they offer.

To measure performance we compute the Receiver Operating Characteristic (ROC) curve, which measures the true 
positive rate against the false positive rate for each dataset given a pair of segment descriptors and their labels.

\figref{fig:roc} presents three ROCs for KITTI, Burgess Field, and RobotCar datasets. The proposed method, \textbf{DSM}, is 
consistently second-best with a difference of 1-2\% --- but again we emphasize the computational efficiency of \textbf{DSM}. 
The classification (Feature Quality) branch is trained specifically to provide additional information about whether two point cloud 
segments match each other. By directly maximizing this metric, the increase in performance of the feature quality branch is 
evident. We note that all methods perform slightly worse on the natural dataset (Burgess Field) as compared to KITTI, as the 
trees and bushes in that scene are more challenging to classify. Similarly, all methods perform slightly better on the RobotCar 
dataset, due to the better alignment and similarity of segments. In line with the literature, we show that learning methods 
outperform methods based on engineered features~\cite{dub2017icra,tinchev18seeing}.

\subsection*{Generalizability}

In this section we study the different models to test their ability to generalize to out-of-domain data. We employ the area 
under the ROC curve (AUC) metric for each permutation of training and testing datasets. We trained our baselines and the 
proposed model to determine how performance degrades when the testing data domain differs from the training data.

\tabref{table:percentage_results} presents the AUC for each of the deep learning and Random Forest models. 
For each testing dataset, the best performing method is indicated with a percentage value of 
100\% (in green). For ease of reading we compute the loss in performance relative to that best method, expressed again as a 
percentage (always less than 100\%). The second-best method is highlighted in 
orange.

We note that the Feature Quality branch for the proposed method generalizes much better than the approach without the 
feature quality classification branch. Regardless of the choice of training and testing dataset, the drop in performance of our 
model is comparable to the other learning approaches. \textbf{DSM} gives consistently the second best performance, trailing 
just behind our previous method ESM~\cite{tinchev2019learning} --- with much reduced computation, as we will later see.

When evaluating the hand-crafted feature methods (NSM, SM-Eig, SM-ESF) on the KITTI dataset, 
we found that Random Forests trained on Oxford RobotCar actually outperform the Random Forest models trained on KITTI. We 
attribute this to the larger sample size of the Oxford RobotCar dataset and the better alignment of the training data. For 
the learning approaches, the performance drop is 10\% in the worst cases. Interestingly, \textbf{DSM-FQ}, trained on 
Burgess Field, performs slightly better than when trained on KITTI. We attribute this small variation to the FQ branch of the 
architecture.

When testing on the Burgess Field dataset, the proposed method (\textbf{DSM} and \textbf{DSM-FQ}) has just a 2\% 
drop of performance when trained on KITTI, and 3\% when trained on RobotCar. This is a generalization similar to 
ESM~\cite{tinchev2019learning}, but better than the other baselines. 

On the RobotCar dataset, the trained models of \textbf{DSM} and \textbf{DSM-FQ} consistently perform second best, by about 
$1\%$ percent --- second to ESM~\cite{tinchev2019learning}. We also note that the scores reported by all algorithms when 
tested on the RobotCar dataset are consistently higher regardless of the training dataset (inline with the ROC experiment).

In summary, the proposed method, \textbf{DSM}, gives performance which is almost directly comparable to the best performing 
model with only a 1-2\% reduction in performance.

\subsection*{Rotation Variation}

So far we showed that deep learning methods perform well when trained and tested on the same datasets, but 
also generalize 
well to unseen data.  In this section we investigate a common issue in point-based networks, such as the one used 
in~\cite{tinchev2019learning}. These types of architectures struggle with viewpoint variation --- when the same scene is sensed 
from different orientations. This is because robustness to this variation is not typically ingrained within the networks.

We train our descriptor, \textbf{DSM}, with rotation augmentation which applies random yaw rotations uniformly sampled 
between $0$ and $180$ degrees for each sample. We also further pre-process the individual clouds before inputting them 
to the 
network. The computational cost associated with this is negligible as we will see later. We chose to compute the following 
distance as a metric to evaluate how robust our model is to random rotations: $\delta=d(c_n,c_0) / Z$, where $d$ is the 
distance in feature space between a segment $c_n$, where $n \in \{0,10,20...360\}$ represents the features of a segment 
rotated every $10$ degrees and $c_0$ (the unrotated reference). $Z$ is a normalization factor that represents the mean distance 
between all other samples in the dataset. Thus, the metric is lower for architectures that are more robust to rotations.

\begin{figure}[t]
	\centering
	\includegraphics[width=1.0\linewidth]{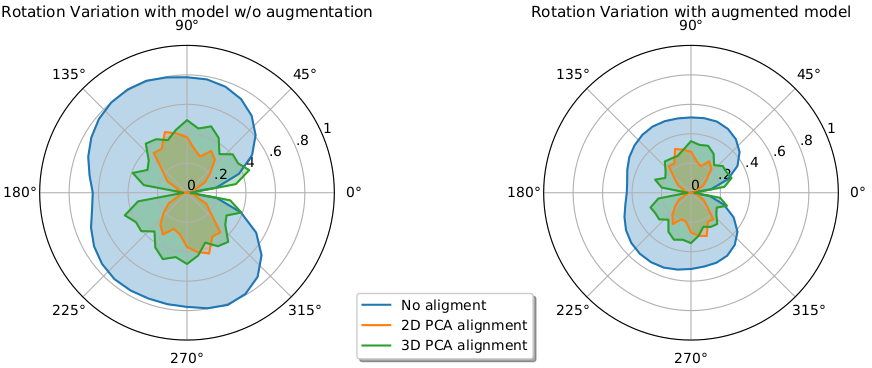}
	\caption{Rotation variation with two models trained without augmentation (left) and with it (right). We have evaluated two 
		alignment strategies based on PCA at different rotation angles. \textbf{Lower score indicates better performance}.}
	\label{fig:rotation_invariance}
	\figvspace
\end{figure}

\figref{fig:rotation_invariance} presents two different variations of our model - with and without applying rotation 
augmentation to random segments during training, as well as 3 different modes of preprocessing: no preprocessing (similar 
to~\cite{tinchev2019learning}), 2D PCA alignment, and 3D PCA alignment. The angle indicates a sample at different rotations 
($n$), while the radius value corresponds to $\delta$. The result indicates that while augmentation alleviates the issue, 
pre-processing is essential for mitigating rotation variations. In experiments which follow we utilize 2D PCA alignment 
preprocessing.



\begin{table*}[t]
	\resizebox{\textwidth}{!}{%
		\centering
		\begin{tabular}{|l|c|c|c|c|c|c|c|c|c|c|} 
			\hline
			\rowcolor[rgb]{1,0.8,0.404} \multicolumn{5}{|c|}{\textbf{Test Set Characteristics}}
			& \multicolumn{1}{l|}{{\cellcolor[rgb]{0.937,0.937,0.937}}\textbf{} } 
			& \multicolumn{5}{c|}{\textbf{Number of Localizations}}
			\\ 
			\hline
			\rowcolor[rgb]{0.937,0.937,0.937} \multicolumn{1}{|c|}{ \textbf{Name} } 
			& \textbf{Type}  
			& \textbf{Length}
			& \textbf{Sensor}        
			& \textbf{Num. Clouds}  
			& \textbf{}                                                           
			& \textbf{SegMatch}  
			& \textbf{NSM}
			& \textbf{SegMap}                               
			& \textbf{ESM}                                
			& \textbf{DSM} 
			\\ 
			\hline
			
			{\cellcolor[rgb]{0.937,0.937,0.937}}\textbf{Oxford RobotCar}                     
			& Urban
			& 50310 m
			& Push-broom
			& 1118
			& \multicolumn{1}{l|}{{\cellcolor[rgb]{0.937,0.937,0.937}}}
			& 80
			& 346
			& {\cellcolor[rgb]{0.604,1,0.6}}\textbf{915}
			& 896
			& 773
			\\
			\hline
			
			{\cellcolor[rgb]{0.937,0.937,0.937}}\textbf{KITTI}                      
			& Urban          
			& 1300 m           
			& Velodyne-HDL-64E 
			& 134                   
			& {\cellcolor[rgb]{0.937,0.937,0.937}}                                
			& 41                 
			& 53               
			& 62                                         
			& {\cellcolor[rgb]{0.604,1,0.6}}\textbf{63}
			& 62
			\\ 
			\hline
			
			{\cellcolor[rgb]{0.937,0.937,0.937}}\textbf{Fire Service College}            
			& Industrial
			& 190 m
			& Velodyne-VLP-16E
			& 169                   
			& \multicolumn{1}{l|}{{\cellcolor[rgb]{0.937,0.937,0.937}}}           
			& 13                   
			& 12                 
			& 35                                           
			& 45
			& {\cellcolor[rgb]{0.604,1,0.6}}\textbf{46}
			\\ 
			\hline
			
			{\cellcolor[rgb]{0.937,0.937,0.937}}\textbf{George Square}              
			& City Park      
			& 500 m            
			& Push-broom             
			& 30                    
			& {\cellcolor[rgb]{0.937,0.937,0.937}}                                
			& 8                  
			& 9                
			& 8                                          
			& {\cellcolor[rgb]{0.604,1,0.6}}\textbf{11}   
			& 10
			\\ 
			\hline
			
			{\cellcolor[rgb]{0.937,0.937,0.937}}\textbf{Cornbury Forest}              
			& Forest         
			& 700 m            
			& Velodyne-HDL-32E        
			& 1125                  
			& {\cellcolor[rgb]{0.937,0.937,0.937}}                                
			& N/A                
			& 29               
			& 231                                        
			& {\cellcolor[rgb]{0.604,1,0.6}}\textbf{248}
			& 208
			\\ 
			\hline
			
			{\cellcolor[rgb]{0.937,0.937,0.937}}\textbf{Burgess Field}              
			& Forest         
			& 2800 m           
			& Push-broom             
			& 130                   
			& {\cellcolor[rgb]{0.937,0.937,0.937}}                                
			& 41                 
			& 54               
			& {\cellcolor[rgb]{0.604,1,0.6}}\textbf{95}  
			& 94                                          
			& 90
			\\ 
			\hline
			
		\end{tabular}
	}
	\caption{Number of localizations across all datasets.}
	\label{tab:loc_results}
	\vspace{-1.5em}
\end{table*}



\begin{table*}[t!]
	\resizebox{\textwidth}{!}{%
		\centering
		
		\begin{tabular}{|l|c|l|c|c|c|c|c|c|c|} 
			\hline
			
			\multicolumn{10}{|c|}
			{{\cellcolor[rgb]{1,0.8,0.404}}\textbf{CPU Multi-core execution times} }       
			\\ 
			\hline
			
			\rowcolor[rgb]{0.937,0.937,0.937} \textbf{Algorithm}  
			& \textbf{Desc. size [dim]}  
			& \textbf{}                                                 
			& \textbf{Segmentation [ms]}              
			& \textbf{Preprocessing [ms]}             
			& \textbf{Descriptor [ms]}                   
			& \textbf{Matching [ms] (K)}                     
			& \textbf{Prunning [ms] (RF)}              
			& \textbf{Pose Est. [ms]}                 
			& \textbf{Total [ms]}   
			\\ 
			\hline
			
			{\cellcolor[rgb]{0.937,0.937,0.937}}SegMatch           
			& 7 (647)              
			& {\cellcolor[rgb]{0.937,0.937,0.937}}                      
			& {\cellcolor[rgb]{0.608,1,0.608}}17 
			& 0                                  
			& {\cellcolor[rgb]{1,0.965,0.608}}605   
			& {\cellcolor[rgb]{0.729,1,0.608}}50 (200)  
			& {\cellcolor[rgb]{1,0.965,0.608}}755 
			& {\cellcolor[rgb]{0.729,1,0.608}}64 
			& 1491
			\\ 
			\hline
			
			{\cellcolor[rgb]{0.937,0.937,0.937}}NSM                
			& 66                   
			& {\cellcolor[rgb]{0.937,0.937,0.937}}                      
			& {\cellcolor[rgb]{0.608,1,0.608}}17 
			& 0                                  
			& {\cellcolor[rgb]{0.608,1,0.608}}24    
			& {\cellcolor[rgb]{0.906,1,0.608}}207 (200) 
			& {\cellcolor[rgb]{1,0.965,0.608}}913 
			& {\cellcolor[rgb]{0.729,1,0.608}}65 
			& 1226
			\\ 
			\hline
			
			{\cellcolor[rgb]{0.937,0.937,0.937}}SegMap             
			& 64                   
			& {\cellcolor[rgb]{0.937,0.937,0.937}}                      
			& {\cellcolor[rgb]{0.608,1,0.608}}17 
			& {\cellcolor[rgb]{0.608,1,0.608}}15 
			& {\cellcolor[rgb]{1,0.745,0.608}}5902  
			& {\cellcolor[rgb]{0.608,1,0.608}}19 (25)   
			& 0                                   
			& {\cellcolor[rgb]{0.608,1,0.608}}21 
			& 5974
			\\ 
			\hline
			
			{\cellcolor[rgb]{0.937,0.937,0.937}}ESM                
			& 16                   
			& {\cellcolor[rgb]{0.937,0.937,0.937}}                      
			& {\cellcolor[rgb]{0.608,1,0.608}}17 
			& {\cellcolor[rgb]{0.608,1,0.608}}3  
			& {\cellcolor[rgb]{1,0.965,0.608}}578   
			& {\cellcolor[rgb]{0.608,1,0.608}}8 (25)    
			& 0                                   
			& {\cellcolor[rgb]{0.608,1,0.608}}9  
			& 615
			\\ 
			\hline
			
			{\cellcolor[rgb]{0.937,0.937,0.937}}DSM                
			& 16                   
			& \multicolumn{1}{c|}{{\cellcolor[rgb]{0.937,0.937,0.937}}} 
			& {\cellcolor[rgb]{0.608,1,0.608}}17 
			& {\cellcolor[rgb]{0.608,1,0.608}}8  
			& {\cellcolor[rgb]{0.906,1,0.608}}184   
			& {\cellcolor[rgb]{0.608,1,0.608}}8 (25)    
			& 0                                   
			& {\cellcolor[rgb]{0.608,1,0.608}}9  
			& 226
			\\ 
			\hline

			\multicolumn{10}{|c|}
			{{\cellcolor[rgb]{1,0.8,0.404}}\textbf{CPU Single-core execution times} }                 
			\\ 
			\hline
			
			\rowcolor[rgb]{0.937,0.937,0.937} \textbf{Algorithm}   
			& \textbf{Desc. size [dim]}  
			& \textbf{}                                                 
			& \textbf{Segmentation [ms]}              
			& \textbf{Preprocessing [ms]}             
			& \textbf{Descriptor [ms]}                   
			& \textbf{Matching [ms] (K)}                     
			& \textbf{Prunning [ms] (RF)}              
			& \textbf{Pose Est. [ms]}                 
			& \textbf{Total [ms]}    
			\\ 
			\hline
			
			{\cellcolor[rgb]{0.937,0.937,0.937}}SegMatch           
			& 7 (647)              
			& {\cellcolor[rgb]{0.937,0.937,0.937}}                      
			& {\cellcolor[rgb]{0.608,1,0.608}}21 
			& 0                                  
			& {\cellcolor[rgb]{1,0.965,0.608}}596   
			& {\cellcolor[rgb]{0.729,1,0.608}}70 (200)  
			& {\cellcolor[rgb]{1,0.965,0.608}}773 
			& {\cellcolor[rgb]{0.729,1,0.608}}64 
			& 1524
			\\ 
			\hline
			
			{\cellcolor[rgb]{0.937,0.937,0.937}}NSM                
			& 66                   
			& {\cellcolor[rgb]{0.937,0.937,0.937}}                      
			& {\cellcolor[rgb]{0.608,1,0.608}}21 
			& 0                                  
			& {\cellcolor[rgb]{0.608,1,0.608}}37    
			& {\cellcolor[rgb]{0.906,1,0.608}}213 (200) 
			& {\cellcolor[rgb]{1,0.965,0.608}}936 
			& {\cellcolor[rgb]{0.729,1,0.608}}65 
			& 1272
			\\ 
			\hline
			
			{\cellcolor[rgb]{0.937,0.937,0.937}}SegMap             
			& 64                   
			& {\cellcolor[rgb]{0.937,0.937,0.937}}                      
			& {\cellcolor[rgb]{0.608,1,0.608}}21 
			& {\cellcolor[rgb]{0.608,1,0.608}}26 
			& {\cellcolor[rgb]{1,0.608,0.608}}25945 
			& {\cellcolor[rgb]{0.608,1,0.608}}26 (25)   
			& 0                                   
			& {\cellcolor[rgb]{0.608,1,0.608}}21 
			& 26039          
			\\ 
			\hline
			
			{\cellcolor[rgb]{0.937,0.937,0.937}}ESM                
			& 16                   
			& {\cellcolor[rgb]{0.937,0.937,0.937}}                      
			& {\cellcolor[rgb]{0.608,1,0.608}}21 
			& {\cellcolor[rgb]{0.608,1,0.608}}3  
			& {\cellcolor[rgb]{1,0.882,0.608}}2126  
			& {\cellcolor[rgb]{0.608,1,0.608}}12 (25)   
			& 0                                   
			& {\cellcolor[rgb]{0.608,1,0.608}}11 
			& 2173    
			\\ 
			\hline
			
			{\cellcolor[rgb]{0.937,0.937,0.937}}DSM                
			& 16                   
			& \multicolumn{1}{c|}{{\cellcolor[rgb]{0.937,0.937,0.937}}} 
			& {\cellcolor[rgb]{0.608,1,0.608}}21 
			& {\cellcolor[rgb]{0.608,1,0.608}}8  
			& {\cellcolor[rgb]{1,0.965,0.608}}731    
			& {\cellcolor[rgb]{0.608,1,0.608}}12 (25)   
			& 0                                   
			& {\cellcolor[rgb]{0.608,1,0.608}}11 
			& 783
			\\ 
			\hline
			
			\multicolumn{10}{|c|}{{\cellcolor[rgb]{1,0.8,0.404}}\textbf{GPU Computation comparison} }
			\\ 
			\hline
			
			\rowcolor[rgb]{0.937,0.937,0.937}  \textbf{Algorithm}  
			& \textbf{Desc. size [dim]}  
			& \textbf{}                                                 
			& \textbf{Segmentation [ms]}              
			& \textbf{Preprocessing[ms]}             
			& \textbf{Descriptor [ms]}                   
			& \textbf{Matching [ms] (K)}                     
			& \textbf{Prunning [ms] (RF)}              
			& \textbf{Pose Est. [ms]}                 
			& \textbf{Total [ms]}   
			\\ 
			\hline
			
			{\cellcolor[rgb]{0.937,0.937,0.937}}SegMap             
			& 64                   
			& \multicolumn{1}{c|}{{\cellcolor[rgb]{0.937,0.937,0.937}}} 
			& {\cellcolor[rgb]{0.608,1,0.608}}17 
			& {\cellcolor[rgb]{0.608,1,0.608}}15 
			& {\cellcolor[rgb]{0.608,1,0.608}}15    
			& {\cellcolor[rgb]{0.608,1,0.608}}19 (25)
			& 0                                   
			& {\cellcolor[rgb]{0.608,1,0.608}}15 
			& 81
			\\ 
			\hline
			
			{\cellcolor[rgb]{0.937,0.937,0.937}}ESM                
			& 16                   
			& \multicolumn{1}{c|}{{\cellcolor[rgb]{0.937,0.937,0.937}}} 
			& {\cellcolor[rgb]{0.608,1,0.608}}17 
			& {\cellcolor[rgb]{0.608,1,0.608}}3  
			& {\cellcolor[rgb]{0.608,1,0.608}}2     
			& {\cellcolor[rgb]{0.608,1,0.608}}8 (25)
			& 0                                   
			& {\cellcolor[rgb]{0.608,1,0.608}}6  
			& 36
			\\ 
			\hline
			
			{\cellcolor[rgb]{0.937,0.937,0.937}}DSM                
			& 16                   
			& \multicolumn{1}{c|}{{\cellcolor[rgb]{0.937,0.937,0.937}}} 
			& {\cellcolor[rgb]{0.608,1,0.608}}17 
			& {\cellcolor[rgb]{0.608,1,0.608}}8  
			& {\cellcolor[rgb]{0.608,1,0.608}}2     
			& {\cellcolor[rgb]{0.608,1,0.608}}8 (25)
			& 0                                   
			& {\cellcolor[rgb]{0.608,1,0.608}}6  
			& 41
			\\
			\hline
			
		\end{tabular}
	}
	\caption{Average computation times in milliseconds recorded per point cloud for a localization query on the Burgess Field 
		dataset.}
	\label{tab:performance}
	\tablevspace
\end{table*}


\subsection*{Localization Results}

Thus far we have focused primarily on the deep learning part of our approach. In this section we evaluate the complete 
performance 
of our method --- how well and how often it localizes the current scene within a large prior map. 

First, we evaluated an urban environment, consisting of four sequences from the RobotCar dataset and sequence 00 from KITTI. 
Second, we have considered different types of natural environments: parkland (George Square), forested meadows (Burgess 
Field), and a heavy forest (Cornbury Forest) --- shown in~\figref{fig:combined_datasets}. Third, we tested the algorithms in an 
industrial setting using the ANYmal quadruped at the 
Fire 
Service College, Gloucestershire. These datasets not only present different environments, but also different sensor variations. 
When evaluating the 
datasets, it should be noted that the total number of clouds does not reflect the number of possible localizations. This is due to 
the fact the robots sometimes traversed parts of the environments missing from the prior map. Given this, we were pleased that 
the approaches are robust and did not compute any false localizations in those areas. For the vegetated datasets, and the 
industrial dataset we used the model trained on the Burgess Field dataset. Therefore, the training and testing data is inherently 
different as it comes from different sensors. The rest of the datasets used their corresponding training data - RobotCar trained 
on another 26 sequences, and KITTI trained on Sequences 05 and 06.

\tabref{tab:loc_results} shows the localization results for all the datasets. The proposed method 
performs commensurately to the other learning approaches --- ESM~\cite{tinchev2019learning} and SegMap~\cite{segmap2018}. 
Qualitative examples of the performance of our algorithms is illustrated in Figure~\ref{fig:combined_datasets} and the video in 
the supplementary material. The supplementary material\footnote[3]{Please see the full collection of videos on 
https://ori.ox.ac.uk/lidar-localization/} also shows the proposed algorithm evaluated on the Newer College 
dataset~\cite{ramezani2020newer}. 

\subsection*{Computational Efficiency}
\label{sec:computational_efficiency}

To this end we demonstrated the performance of our method, in this section we study the computational cost associated with 
each method.

We tested the methods on a mobile Intel Xeon E3-1535M CPU with the following configurations: 1) a single-core, 2) multi-cores 
of the same processor, and 3) the CPU with a NVIDIA Titan Xp GPU. We have processed the push-broom point clouds from 
Burgess Field and recorded the mean computation time as milliseconds per localization 
query. 

To provide a fair comparison, we have used the same Euclidean segmentation for the approaches to create the same segments, 
and thus the segmentation time is equivalent. We empirically evaluated the number of neighbors in feature 
space ($K$) to retrieve the most positive localizations, while keeping zero false positives - $25$ for learned approaches, and 
$200$ for hand-engineered ones. The value of $K$ influences the pose estimation stage as well. We also note that the 
descriptor size affects the time for matching and pruning.

\tabref{tab:performance} summarizes the recorded GPU and CPU times for all the approaches. The proposed method, 
\textbf{DSM}, performs three times faster than the previous best~\cite{tinchev2019learning} on a CPU, while performing similarly 
on a GPU. We attribute this to the sampling performed at each layer, as the number of parameters is fairly similar between the 
two architectures - $280$\,K vs $300$\,K. We also note that the cost of preprocessing using PCA alignment is negligible 
compared to the other components and is justified given the benefits we saw from the Rotation Variation experiment. When 
training \textbf{DSM} we 
noticed that the GPU memory consumption compared to ESM~\cite{tinchev2019learning} for the same batch size was 70\% less. 
Therefore, to fully utilize the GPU we have increased the batch size when training \textbf{DSM}. We also note that the pose 
estimator computes a transformation estimate quicker, due to the probability information extracted from the classification 
branch and the use of the 
PROSAC estimator~\cite{prosac2005}.

\section{CONCLUSION}

In this paper we discussed the problem of place recognition based on LIDAR segment matching in both urban and natural 
environments. Our newly proposed method (\textbf{DSM}) works at three times the frequency of our previous best, while 
requiring only a CPU. We showed that by down-sampling in the inner layers of the architecture, we gain this significant 
computation cost reduction. We presented an efficient CPU implementation of the FPS algorithm and we also addressed a 
common issue with rotation variation through experimental comparison and data augmentation. We have thoroughly validated 
our methodology on nine scenarios from six different datasets in various environments. Finally, a significant contribution to the 
Oxford RobotCar dataset was made, and is being made available to the community through this paper, with improved alignment 
and ground truth labeling. We hope that the robotics community will adopt this dataset for more learning tasks.

\small
\bibliographystyle{IEEEtran}
\small
\bibliography{main}
\end{document}